\documentclass[11pt]{article}

\usepackage[dvips]{graphicx}
\usepackage{amsmath,amsthm,amscd}
\usepackage{amssymb}
\usepackage{fancybox}
\usepackage{latexsym}
\usepackage[dvips]{color}

\newtheorem{theorem}{Theorem}
\newtheorem{corollary}[theorem]{Corollary}
\newtheorem{assumption}{Assumption}
\newtheorem{remark}{Remark}
\newtheorem{example}{Example}

\def\pnu{p_{\mathrm{n}}}
\def\pde{p_{\mathrm{d}}}
\def\xnu{x^{(\mathrm{n})}}
\def\xde{x^{(\mathrm{d})}}
\def\nnu{{m_{\mathrm{n}}}}
\def\nde{{m_{\mathrm{d}}}}
\def\mnu{{m_{\mathrm{n}}}}
\def\mde{{m_{\mathrm{d}}}}
\def\fnu{{f_{\mathrm{n}}}}
\def\fde{{f_{\mathrm{d}}}}
\def\iid{{\it i.i.d.\,}}
\def\simiid{{\sim_\iid}}
\def\Real{\mathbb{R}}
\def\Enu{{\mathrm{E}_\mathrm{n}}}
\def\Ede{{\mathrm{E}_\mathrm{d}}}
\def\Pbb{\mathbb{P}}
\def\Ebb{\mathbb{E}}
\def\Vbb{\mathbb{V}}
\def\Vnu{{\mathrm{V}_\mathrm{n}}}
\def\Vnuh{{{\widehat{V}}_\mathrm{n}}}
\def\Vde{{\mathrm{V}_\mathrm{d}}}
\def\Cov{{\mathrm{Cov}}}
\def\cd{\stackrel{d}{\longrightarrow}}
\def\cp{\stackrel{p}{\longrightarrow}}

\title{$f$-divergence estimation and two-sample homogeneity test under semiparametric
density-ratio models}

\author{
  Takafumi Kanamori\\ Nagoya University \\ \tt{kanamori@is.nagoya-u.ac.jp}
  \and
  Taiji Suzuki\\ University of Tokyo\\ \tt{s-taiji@stat.t.u-tokyo.ac.jp}
  \and  
  Masashi Sugiyama\\
  Tokyo Institute of Technology\\ \tt{sugi@cs.titech.ac.jp}
 }

\date{}

\begin{document}
\maketitle
 
\begin{abstract}
 A density ratio is defined by the ratio of two probability densities. We study the
 inference problem of density ratios and apply a semi-parametric 
 density-ratio estimator to the two-sample homogeneity test.
 In the proposed test procedure, the
 $f$-divergence between two probability densities is estimated using 
 a density-ratio estimator. The $f$-divergence estimator 
 is then exploited for the two-sample homogeneity test. 
 We derive the optimal estimator of $f$-divergence in the sense of the asymptotic
 variance, and then investigate the relation between the proposed test procedure and the
 existing score test based on empirical likelihood estimator. 
 Through numerical studies, we illustrate the adequacy of the asymptotic theory for
 finite-sample inference. 
\end{abstract}

\section{Introduction}
\label{sec:Introduction}
In this paper, we study the two-sample homogeneity test under semiparametric density-ratio 
models. The estimator of the density ratio is exploited to obtain a test statistic. 
For two probability densities, $\pnu(x)$ and $\pde(x)$, over a probability space
$\mathcal{X}$, the density ratio $r(x)$ is defined as the ratio of these densities, that
is, 
\[
r(x)=\frac{\pnu(x)}{\pde(x)},
\]
in which $\pnu$ ($\pde$) denotes the ``numerator''
(``denominator'') of the density ratio. 
For statistical examples and motivations of the density ration model, 
see Qin \cite{Biometrika:Qin:1998}, Cox and Ferry \cite{cox91:_robus_logis_discr}
and Kay and Little \cite{kay87:_trans_of_explan_variab_in} and the references
therein. Qin \cite{Biometrika:Qin:1998} has studied the inference of the density ratio
under retrospective sampling plans, and proved that the estimating function obtained from
the prospective likelihood is optimal in a class of unbiased estimating functions under
the semiparametric density ratio models. 
As a similar approach, Cheng and Chu \cite{Bernoulli:Cheng+Chu:2004} have studied a
semiparametric density ratio estimator based on logistic regression. 

The density ratio is closely related to the inference of divergences. 
The divergence is a discrepancy measure between pairs of multivariate probability 
densities, and the $f$-divergence 
\cite{JRSS-B:Ali+Silvey:1966,SSM-Hungary:Csiszar:1967}
is a class of divergences based on the ratio of two
probability densities. 
For a strictly convex function $f$ satisfying $f(1)=0$, the $f$-divergence between 
two probability densities $\pde(x)$ and $\pnu(x)$ is defined by 
\begin{align}
 D_f(\pde,\pnu)
 =\int_\mathcal{X}\pde(x)f\left(\frac{\pnu(x)}{\pde(x)}\right)dx. 
 \label{eqn:f-divergence}
\end{align}
Since $f$ is strictly convex, the $f$-divergence is non-negative and takes zero if and
only if $\pnu=\pde$ holds. 
Popular divergences such as Kullback-Leibler (KL) divergence
\cite{Annals-Math-Stat:Kullback+Leibler:1951}, 
Hellinger distance, and Pearson divergence are included
in the $f$-divergence class. 
In statistics, machine learning, and information theory, the $f$-divergence is often
exploited as a metric between probability distributions, even though the divergence does
not necessarily satisfy the definition of the metric. 

A central topic in this line of research 
is to estimate the divergence based on samples from each 
probability distribution.  
A typical approach is to exploit non-parametric estimators
of the probability densities, 
and then estimate, say, KL-divergence based on the estimated probability densities
\cite{wang06:_neares_neigh_approac_to_estim}. 

In order to estimate the $f$-divergence between two probabilities, 
Keziou \cite{keziou03:_dual_repres_of_phi_diver_and_applic} has exploited the conjugate
expression of the $f$-divergence. Based on the conjugate expression, Keziou and
Leoni-Aubin \cite{kezioua05:_test_of_homog_in_semip}, and Broniatowski and Keziou
\cite{broniatowski09:_param_estim_and_tests_throug} have developed $f$-divergence
estimators for semiparametric density-ratio models. 
Keziou and Leoni-Aubin \cite{kezioua05:_test_of_homog_in_semip} have applied the
$f$-divergence estimator to the one-sample test. 
Recently, Nguyen et al.~\cite{NIPS:Nguyen+etal:2008} has developed a kernel-based
estimator of the $f$-divergence using a non-parametric density-ratio model. 

Once the divergence between two probability densities is estimated, 
the homogeneity test can be conducted. 
In the homogeneity test, the null hypothesis is represented as $H_0:\pnu=\pde$ against the
complementary alternative $H_1:\pnu\neq\pde$. 
If an estimate of $D_f(\pde,\pnu)$ is beyond some positive value, 
the null hypothesis is rejected and the alternative is accepted. 
Keziou \cite{kezioua05:_test_of_homog_in_semip} has studied the homogeneity test 
using $f$-divergence estimator for semiparametric density-ratio models. 
On the other hand, Fokianos et al.~\cite{fokianos01:_semip_approac_to_one_way_layout}
adopted a more direct approach. 
They have proposed the score test derived from the empirical likelihood estimator of
density ratios. 
In our paper, we consider the optimality of $f$-divergence estimators, and investigate the
relation between the test statistic using the $f$-divergence estimator and the score test
derived from the empirical likelihood estimator. 


The rest of this paper is organized as follows: 
In Section~\ref{section:Estimation_of_density_ratio} we introduce unbiased estimators 
of density ratios for semiparametric density-ratio models. 
We also define some notation which is used throughout this paper. 
In Section~\ref{sec:Asymptotics_Estimation_f-Divergence}, we consider the asymptotics of
an $f$-divergence estimator. 
The main results of this paper are presented in
Section~\ref{sec:Optimal_Estimator_f-divergence}
and Section~\ref{sec:Higher-order_Asymptotics}. 
We present the optimal estimator for the $f$-divergence, which is then
exploit for two-sample homogeneity test. 
Broniatowski and Keziou \cite{broniatowski09:_param_estim_and_tests_throug} proposed the
estimator exploiting the conjugate expression of the $f$-divergence, but they argued
neither its optimality nor its efficiency.
A main contribution of this paper is to present the optimal estimator of the
$f$-divergence in the sense of asymptotic variance under the semiparametric density-ratio
models. 
Then, we propose a test statistic based on the optimal $f$-divergence estimator,
and investigate its power function. 
Numerical studies are provided in Section~\ref{sec:Numerical_Studies},
illustrating the adequacy of our asymptotic theory for finite-sample inference.
Section~\ref{section:Conclusion} is devoted to concluding remarks. 
Some calculations are deferred to Appendix.

\section{Estimation of density ratio}
\label{section:Estimation_of_density_ratio}
We introduce the method of estimating density ratios according to Qin
\cite{Biometrika:Qin:1998}. 
Let $\pnu(x)$ and $\pde(x)$ be two
probability densities on probability space $\mathcal{X}$.
Their density ratio is defined as 
\begin{align*}
 r(x)=\frac{\pnu(x)}{\pde(x)}
\end{align*}
for $x\in\mathcal{X}$. Two sets of samples are independently generated from each
probability: 
\begin{align*}
 \xnu_1,\ldots, \xnu_{\nnu}~\simiid~ \pnu,\qquad 
 \xde_1,\ldots, \xde_{\nde}~\simiid~ \pde. 
\end{align*}
The model for the density ratio is defined by $r(x;\theta)$ with the parameter
$\theta\in\Theta\subset\Real^d$. We assume that the true density ratio is represented as 
\begin{align*}
 r(x)=\frac{\pnu(x)}{\pde(x)}=r(x;\theta^*) 
\end{align*}
with some $\theta^*\in\Theta$. 
The model for the density ratio $r(x;\theta)$ is regarded as a semiparametric model for
probability densities. That is, 
even if $r(x;\theta^*)=\pnu(x)/\pde(x)$ is specified, there are yet infinite degrees of
freedom for the probability densities $\pnu$ and $\pde$. 

The moment matching estimator for the density ratio has been proposed by Qin 
\cite{Biometrika:Qin:1998}. Let $\eta(x;\theta)\in\Real^d$ be a vector-valued function
from $\mathcal{X}\times \Theta$ to $\Real^d$, and the estimation function $Q_\eta$ is
defined as 
\begin{align*}
 Q_\eta(\theta)
 &:=
 \frac{1}{\mde}\sum_{i=1}^{\mde}r(\xde_i;\theta)\eta(\xde_i;\theta)
 - \frac{1}{\mnu}\sum_{j=1}^{\mnu}\eta(\xnu_j;\theta). 
\end{align*}
Since $\pnu(x)=r(x;\theta^*)\pde(x)$ holds, the expectation of $Q_\eta(\theta)$ over the
observed samples vanishes at $\theta=\theta^*$. In addition, the estimation function
$Q_\eta(\theta)$ converges to its expectation in the large sample limit. 
Thus, the estimator $\widehat{\theta}$ defined as 
a solution of the estimating equation 
\begin{align*}
 Q_\eta(\widehat{\theta})=0
\end{align*}
has the statistical consistency under the mild  assumption, see \cite{Biometrika:Qin:1998}
for details. 

The moment matching estimation of the density ratio contains a wide range of
estimators. 
Several authors such as Nguyen et al.~\cite{NIPS:Nguyen+etal:2008}, Keziou and 
Leoni-Aubin \cite{kezioua05:_test_of_homog_in_semip}, Sugiyama et
al.~\cite{NIPS:Sugiyama+etal:2008} 
and Kanamori et al.~\cite{t.09:_effic_direc_densit_ratio_estim}
have proposed various density-ratio estimators.
These estimators with a finite-dimensional model $r(x;\theta)$ can all be
represented as a moment matching estimator. 
These existing methods, however, are intended to be applied with kernel methods which have
been developed in machine learning  \cite{book:Schoelkopf+Smola:2002,book:Wahba:1990}. 
As another approach to density ratio estimation, 
Kwik and Mielniczuk \cite{kwik89:_estim_densit_ratio_with_applic}, 
Jacoba and Oliveirab \cite{jacoba97:_kernel_estim_of_gener_radon_nikod_deriv}, 
and Bensaid and Fabre \cite{bensaid07:_optim_asymp_quadr_error_of}
have exploited the kernel density estimator, and studied convergence properties of
estimators under several theoretical assumptions. 

Before we present the asymptotic results, we prepare some notation. 
Let $N_d(\mu,\Sigma)$ be the $d$-dimensional normal distribution 
with the mean vector $\mu$ and the variance-covariance matrix $\Sigma$. 
The dimension $d$ may be dropped if there is no confusion. 
$\Enu[\,\cdot\,]$ and $\Vnu[\,\cdot\,]$ denote the expectation and the variance 
(or the variance-covariance matrix for multi-dimensional random variables) under the 
probability $\pnu$, and $\Ede[\,\cdot\,]$ and $\Vde[\,\cdot\,]$ are defined in the same way for
the probability $\pde$. 
The expectation and the variance under all samples, 
$\xnu_i\, (i=1,\ldots,\mnu),\ \xde_j\,(j=1,\ldots,\mde)$ are denoted as 
$\Ebb[\,\cdot\,]$ and $\Vbb[\,\cdot\,]$, respectively. 
The covariance matrix between two random variables under all samples are also denoted as 
$\Cov[\,\cdot,\cdot\,]$. 
The first and the second derivative of the function $f:\Real\rightarrow\Real$ are denotes
as $f'$ and $f''$, respectively. 
Let $\partial_i$ be the partial differential operator with respect to the parameter
$\theta$, that is $\partial_i=\frac{\partial}{\partial\theta_i}$. 
The gradient column vector of the function $g$ with respect to the parameter $\theta$ is
denoted as $\nabla g$, i.e., $\nabla g=(\partial_1g,\ldots,\partial_dg)^T$. 
For a vector-valued function $\eta(x;\theta)=(\eta_1(x;\theta),\ldots,\eta_d(x;\theta))$, 
let $\mathcal{L}[\eta(x;\theta)]$ be the linear space 
\begin{align*}
 \mathcal{L}[\eta(x;\theta)]:=
 \bigg\{\,
 \sum_{k=1}^{d} a_k\,\eta_k(x;\theta)~
 \big|~a_1,\ldots,a_d\in\Real \,\bigg\}. 
\end{align*}
In this paper, the linear space $\mathcal{L}[\nabla\log r(x;\theta)]$ defined by
\begin{align*}
 \mathcal{L}[\nabla\log r(x;\theta)]:=
 \bigg\{\,\sum_{k=1}^{d} a_k\,\partial_k\log{r}(x;\theta)~
 \big|~a_1,\ldots,a_d\in\Real \,\bigg\}
\end{align*}
plays the central role. 

We introduce the asymptotics of density ratio estimation. 
Let $\rho$ and $m$ be
\begin{align*}
 \rho:=\frac{\mnu}{\mde},\qquad 
 m:=\bigg(\frac{1}{\mnu}+\frac{1}{\mde}\bigg)^{-1}
 =\frac{\mnu\mde}{\mnu+\mde}, 
\end{align*}
respectively, and the $d$ by $d$ matrix $U_\eta$ be  
\begin{align*}
U_\eta=\Enu[\eta(x;\theta)\nabla \log r(x;\theta)^T ], 
\end{align*}
where $\eta(x;\theta)$ is a $d$-dimensional vector-valued function. 
Suppose that $U_\eta$ is non-degenerate in the vicinity of $\theta=\theta^*$. 
Below, the notation $\rho=\mnu/\mde$ is also used as the large sample limit of
$\mnu/\mde$, and we assume that $0<\rho<\infty$ holds even in the limit. 
The asymptotic expansion of the estimating equation $Q_\eta(\widehat{\theta})=0$ around
$\theta=\theta^*$ yields the following convergence in law, 
\begin{align}
 \sqrt{m}(\widehat{\theta}-\theta^*)=-\sqrt{m}U_\eta^{-1}Q_\eta+o_p(1) \cd 
 N_d\bigg( 0,\,U_\eta^{-1}\frac{\rho \Vde[r\eta]+\Vnu[\eta]}{\rho+1}(U_\eta^T)^{-1} \bigg), 
 \label{eqn:asymptotics-parameter}
\end{align}
in which $\theta$ is set to $\theta^*$. 
The formula above is derived from the equalities 
\begin{align*}
 \mathbb{E}[Q_\eta]=0,\qquad 
 m\cdot \mathbb{E}[Q_\eta Q_{\eta}^T]
 =
 \frac{\rho\Vde[r\eta]+\Vnu[\eta] }{\rho+1}. 
\end{align*}
Qin \cite{Biometrika:Qin:1998} has shown that the prospective likelihood minimizes the
asymptotic variance in the class of moment matching estimators. 
More precisely, for the density ratio model
\begin{align}
 r(x;\theta)=\exp\{\alpha+\phi(x;\beta)\},\qquad
 \theta=(\alpha,\beta)\in\Real\times\Real^{d-1}, 
 \label{eqn:Qin_ratio-model}
\end{align}
the vector-valued function $\eta_{\mathrm{opt}}$ defined by 
\begin{align}
 \eta_{\mathrm{opt}}(x;\theta)=\frac{1}{1+\rho r(x;\theta)}\nabla\log r(x;\theta)
 \label{eqn:optimal-estimator}
\end{align}
minimizes the asymptotic variance of \eqref{eqn:asymptotics-parameter}.

\section{Estimation of $f$-divergence}
\label{sec:Asymptotics_Estimation_f-Divergence}
We consider the estimation of $f$-divergence. 
As shown in \eqref{eqn:f-divergence}, the $f$-divergence is represented as the expectation
of the transformed density ratio $f(r(x))$, that is, 
\begin{align*}
 D_f(\pde,\pnu)
 =\int\pde(x)f\left(\frac{\pnu(x)}{\pde(x)}\right)dx
 =\int\pde(x)f\big(r(x)\big)dx,
\end{align*}
for $r(x)=\pnu(x)/\pde(x)$. 
Once the density ratio is estimated by $r(x;\widehat{\theta})$, 
the $f$-divergence is also estimated by the empirical mean of $f(r(x;\widehat{\theta}))$ 
over the samples from $\pde$. Here we consider an extended estimator. 
Suppose that the convex function $f$ is decomposed into two terms, 
\begin{align}
 f(r)=\fde(r)+r\fnu(r).
 \label{eqn:f-docomp}
\end{align}
Then the $f$-divergence is represented as
\begin{align}
 \int\pde(x)f\big(r(x)\big)dx
 =
 \int\pde(x)\fde(r(x))dx+\int\pnu(x)\fnu(r(x))dx
 \label{eqn:decomposed-fdiv}
\end{align}
since $r(x)=\pnu(x)/\pde(x)$ holds. 
Note that the decomposition \eqref{eqn:f-docomp} includes the conjugate representation
$f(r)=-f^*(f'(r))+rf'(r)$ with the conjugate function $f^*$ 
\cite{rockafellar70:_convex_analy}. 
Keziou \cite{keziou03:_dual_repres_of_phi_diver_and_applic} has exploited the conjugate
representation for the estimation of the $f$-divergence. 
The empirical variant of \eqref{eqn:decomposed-fdiv} provides an estimate of the
$f$-divergence, 
\begin{align}
 \widehat{D}_f=
 \frac{1}{\mde}\sum_{i=1}^{\mde}\fde(r(\xde_i;\widehat{\theta}))+
 \frac{1}{\mnu}\sum_{j=1}^{\mnu}\fnu(r(\xnu_j;\widehat{\theta})), 
\label{eqn:estimator-f-divergence}
\end{align}
where the parameter $\widehat{\theta}$ is estimated by the estimation function $Q_\eta$. 
Using the estimator $\widehat{D}_f$, we can conduct the homogeneity test with hypotheses 
\begin{align}
 H_0: \pnu=\pde,\qquad  H_1: \pnu\neq \pde. 
 \label{eqn:homogeneity_test}
\end{align}
When the null hypothesis is true, the $f$-divergence $D_f(\pde,\pnu)$ is equal to zero and
otherwise $D_f(\pde,\pnu)$ takes a positive real value. Thus, the null hypothesis will be
rejected when $\widehat{D}_f>t$ holds, where $t$ is a positive constant determined from
the significance level of the test. 

We consider the statistical properties of the estimator $\widehat{D}_f$. 
The estimator 
\eqref{eqn:estimator-f-divergence}
depends on two choices: one is
the vector-valued function $\eta$ for the estimation of the density ratio, and 
the other is the decomposition of $f$, i.e., $\fde$ and $\fnu$. 
For the decomposition $f(r)=\fde(r)+r\fnu(r)$, let us define 
\begin{align*}
 \Pbb{f}
 &:= 
 \sqrt{\frac{\rho}{\rho+1}}
 \frac{1}{\sqrt{\mde}}\sum_{i=1}^{\mde} \big[\fde(r(\xde_i;\theta)-\Ede[\fde(r(x;\theta))]\big]\\
 &\phantom{:=}
 +\sqrt{\frac{1}{\rho+1}}
 \frac{1}{\sqrt{\mnu}}\sum_{j=1}^{\mnu}\big[\fnu(r(\xnu_j;\theta)-\Enu[\fnu(r(x;\theta))]\big], 
\end{align*}
and the $d$-dimensional vector $c\in\Real^d$ be 
\begin{align*}
 c  &:= \Enu\big[\{f'(r(x;\theta))-\fnu(r(x;\theta))\}\nabla\log r(x;\theta)\big]. 
\end{align*}
Then, the first order asymptotic expansion of $\widehat{D}_f$ with $f(r)=\fde(r)+r\fnu(r)$
yields that 
\begin{align}
 \sqrt{m}(\widehat{D}_f-D_f)
 &=\Pbb{f}-\sqrt{m}\,c^T U_\eta^{-1}Q_\eta+o_p(1), 
 \label{eqn:f-div-asympto}
\end{align}
in which $D_f$ denotes $D_f(\pde,\pnu)$ and the functions are evaluated at
$\theta=\theta^*$. Based on the above formula, we derive the estimator attaining the
minimum asymptotic variance of \eqref{eqn:f-div-asympto}.

\section{Optimal Estimator of $f$-divergence}
\label{sec:Optimal_Estimator_f-divergence} 
We consider the optimal estimator of the $f$-divergence in the sense of the asymptotic
variance. Some assumptions to be imposed are shown below. 
\begin{assumption}
 \label{cond:model-assumption}
 The density ratio model $r(x;\theta)$ and the function $f$ of the $f$-divergence
 satisfy the following conditions. 
 \begin{description}
  \item[{\rm (a)}] The model $r(x;\theta)$ includes the constant function $1$. 
  \item[{\rm (b)}] For any $\theta\in\Theta$, 
	$1\in\mathcal{L}[\nabla\log{r}(x;\theta)]$ holds. 
  \item[{\rm (c)}] $f(1)=f'(1)=0$. 
 \end{description}
\end{assumption}
As shown in Remark~\ref{remark:1inLinearSpace} below, standard models of density ratios
satisfy (a) and (b) of Assumption~\ref{cond:model-assumption}. 
\begin{remark}
\label{remark:1inLinearSpace}
 Let $\phi(x)=(\phi_1(x),\ldots,\phi_d(x))^T\in\Real^d$ 
 be a vector-valued function defined on $\mathcal{X}$ such that $\phi_1(x)=1$. 
 The exponential model
 $r(x;\theta)=\exp\{\theta^T\phi(x)\}$, that is, the model \eqref{eqn:Qin_ratio-model}
 satisfies (a) and (b) in Assumption \ref{cond:model-assumption}. 
 In the same way, we see that the linear model $r(x;\theta)=\theta^T\phi(x)$ also meets
 the conditions. Indeed, the linear space $\mathcal{L}[\nabla\log r(x;\theta)]$ is spanned
 by $\{\phi_1/r,\ldots,\phi_d/r\}$ and the equality $\theta^T \phi/r(x;\theta)=1$ holds
 for all $\theta\in\Theta$. 
\end{remark}

We compare the asymptotic variance of two estimators for the $f$-divergence; 
one is the estimator $\widehat{D}_f$ derived from the moment matching estimator using
$\eta(x;\theta)$ and the decomposition $f(r)=\fde(r)+r\fnu(r)$, and the other is 
the estimator $\bar{D}_f$ defined by the density ratio estimator $\bar{\eta}(x;\theta)$
and the decomposition $f(r)=\bar{\fde}(r)+r\bar{\fnu}(r)$. 
For each estimator, the asymptotic expansion of $\widehat{D}_f$ is given as
\begin{align*}
 \sqrt{m}(\widehat{D}_f-D_f)=\Pbb{f}-\sqrt{m}\,c^T U_\eta^{-1}Q_\eta+o_p(1)
\end{align*}
and 
\begin{align*}
 \sqrt{m}(\bar{D}_f-D_f)=
 \bar{\Pbb}{f}-\sqrt{m}\,\bar{c}^T U_{\bar{\eta}}^{-1}Q_{\bar{\eta}}+o_p(1), 
\end{align*}
respectively, where $\bar{\Pbb}{f}$ and $\bar{c}$ are defined by
\begin{align*}
 \bar{\Pbb}{f}
 &:= 
  \sqrt{\frac{\rho}{\rho+1}}
 \frac{1}{\sqrt{\mde}}\sum_{i=1}^{\mde} \big[\bar{\fde}(r(\xde_i;\theta)-\Ede[\bar{\fde}(r(x;\theta))]\big]\\
 &\phantom{:=}
 +\sqrt{\frac{1}{\rho+1}}
 \frac{1}{\sqrt{\mnu}}\sum_{j=1}^{\mnu}
 \big[\bar{\fnu}(r(\xnu_j;\theta)-\Enu[\bar{\fnu}(r(x;\theta))]\big], \\
 \bar{c} 
 &:=
 \Enu[\{f'(r(x;\theta))-\bar{\fnu}(r(x;\theta))\}\nabla\log r(x;\theta)], 
\end{align*}
and the functions are evaluated at $\theta=\theta^*$. 
In order to compare the variances of these estimators, we consider the following
inequality, 
\begin{align*}
 0 ~\leq~ \mathbb{V}[\widehat{D}_f-\bar{D}_f]
 ~=~
 \mathbb{V}[\widehat{D}_f]-\mathbb{V}[\bar{D}_f]
 -2\,\mathrm{Cov}[\,\widehat{D}_f-\bar{D}_f,\,\bar{D}_f\,].
\end{align*}
Suppose that the covariance above vanishes for any $\widehat{D}_f$.
Then we have the inequality
\begin{align*}
 \mathbb{V}[\bar{D}_f] \leq  \mathbb{V}[\widehat{D}_f]
\end{align*}
This implies that the estimator $\bar{D}_f$ is the asymptotically optimal estimator for
the $f$-divergence. 

Under Assumption \ref{cond:model-assumption}, 
some calculation of the covariance yields the equality
\begin{align}
 &\phantom{=}
 \mde\,\mathrm{Cov}[\,\widehat{D}_f-\bar{D}_f,\bar{D}_f\,]
 \nonumber\\
 &=
 \Enu\big[
 \big\{\bar{\fnu}(r)-\fnu(r)
 +\bar{c}^T U_{\bar{\eta}}^{-1}\bar{\eta}-c^T U_{\eta}^{-1}\eta\big\}
 \big\{f(r)-(r+\rho^{-1})(\bar{\fnu}(r)+\bar{c}^T U_{\bar{\eta}}^{-1}\bar{\eta})\big\}
\big], 
 \label{eqn:covariance-computation}
\end{align}
in which $r$ denotes the density ratio $r(x)$ and the functions are evaluated at
$\theta=\theta^*$. We study the sufficient condition that the above covariance vanishes. 
\begin{theorem}
 \label{theorem:easy_case}
 Under Assumption \ref{cond:model-assumption}, suppose
 that $\bar{\fde}(r(x;\theta))$, $\bar{\fnu}(r(x;\theta))$, and $\bar{\eta}(x;\theta)$ satisfy
 \begin{align}
  f(r(x;\theta))-(r(x;\theta)+\rho^{-1})(\bar{\fnu}(r(x;\theta))+
  \bar{c}^T U_{\bar{\eta}}^{-1}\bar{\eta}(x;\theta))
  \in\mathcal{L}[\nabla\log r(x;\theta)]
  \label{eqn:theorem_easy_case_condition}
 \end{align}
 for all $\theta\in\Theta$. Then the estimator 
 $\bar{D}_f$
 using $\bar{\eta}$ and the decomposition $f(r)=\bar{\fde}(r)+r\bar{\fnu}(r)$ 
 uniformly attains the minimum asymptotic variance. 
\end{theorem}
\begin{proof}
 For any $\pnu$ and $\pde$ such that $\pnu(x)/\pde(x)=r(x;\theta)$, we have 
 \begin{align*}
  &\phantom{=}\Enu\big[
  \big\{\bar{\fnu}(r(x;\theta))-\fnu(r(x;\theta))+\bar{c}^T U_{\bar{\eta}}^{-1}\bar{\eta}(x;\theta)
  -c^T U_{f}^{-1}\eta(x;\theta)\big\}\nabla\log r(x;\theta)^T\big]\\
  &=
  \Enu\big[(\bar{\fnu}(r(x;\theta))-\fnu(r(x;\theta)))\nabla\log r(x;\theta)^T\big]+\bar{c}^T-c^T\\
  &=0. 
 \end{align*} 
 Hence, when \eqref{eqn:theorem_easy_case_condition} holds, the covariance
 \eqref{eqn:covariance-computation} vanishes for any $\eta$ and any 
 decomposition of $f$. 
\end{proof}
Clearly, any optimal estimator of the $f$-divergence achieves the same asymptotic
variance. In the following corollaries, we present some sufficient conditions of 
\eqref{eqn:theorem_easy_case_condition}. 
\begin{corollary}
 \label{corollary:QinEstimator_optimal}
 Under Assumption \ref{cond:model-assumption}, suppose that 
 \begin{align}
  f(r(x;\theta))-(r(x;\theta)+\rho^{-1})\bar{\fnu}(r(x;\theta))
  \in\mathcal{L}[\nabla\log r(x;\theta)]
  \label{eqn:cor_QinEstimator_Opt}
 \end{align} 
 holds for all $\theta\in\Theta$. Then, the function $\bar{\eta}=\eta_{\mathrm{opt}}$
 defined in \eqref{eqn:optimal-estimator} with the decomposition 
 $\bar{\fde}(r),\bar{\fnu}(r)$ satisfies the condition 
 \eqref{eqn:theorem_easy_case_condition}. 
\end{corollary}
\begin{proof}
 For $r(x;\theta)$, $\eta_{\mathrm{opt}}(x;\theta)$, and $\bar{\fnu}$, we have
 \begin{align*}
  &\phantom{=}
  f(r(x;\theta))-(r(x;\theta)+\rho^{-1})(\bar{\fnu}(r(x;\theta))+\bar{c}^T 
  U_{\eta_\mathrm{opt}}^{-1}\eta_\mathrm{opt}(x;\theta)) \\
  &= 
  f(r(x;\theta))-(r(x;\theta)+\rho^{-1})\bar{\fnu}(r(x;\theta))
  -\rho^{-1}\bar{c}^T U_{\eta_\mathrm{opt}}^{-1}\nabla\log r(x;\theta). 
 \end{align*}
 Under the condition \eqref{eqn:cor_QinEstimator_Opt}, 
 we see that the above expression is included in the linear space 
 $\mathcal{L}[\nabla\log{r}(x;\theta)]$. 
\end{proof}
Based on Corollary \ref{corollary:QinEstimator_optimal}
we see that the estimator defined from 
\begin{align}
 \fde(r)=\frac{f(r)}{1+\rho r},\quad
 \fnu(r)=\frac{\rho f(r)}{1+\rho r},\quad 
 \text{and}\quad
 \eta(x;\theta)=\eta_{\mathrm{opt}}(x;\theta)
 =\frac{1}{1+\rho r(x;\theta)}\nabla\log r(x;\theta)
 \label{eqn:optimal-estimator-with-QinEstimator}
\end{align}
leads to an optimal estimator of the $f$-divergence. 

We show another sufficient condition. 
\begin{corollary}
 \label{corollary:other_optimal_condition}
 Under Assumption \ref{cond:model-assumption}, suppose that 
 \begin{align*}
  \begin{array}{l}
   \displaystyle
    \phantom{\text{and}}\quad
   f(r(x;\theta))-(r(x;\theta)+\rho^{-1})f'(r(x;\theta))\in
   \mathcal{L}[\nabla\log{r}(x;\theta)],\vspace*{2mm}\\
   \text{and}\quad
   \displaystyle
  f'(r(x;\theta))-\bar{\fnu}(r(x;\theta))\in\mathcal{L}[\bar{\eta}(x;\theta)]
  \end{array}
 \end{align*} 
 hold for all $\theta\in\Theta$. Then the decomposition $f(r)=\bar{\fde}(r)+r\bar{\fnu}(r)$
 and the vector-valued function $\bar{\eta}(x;\theta)$ satisfy
 \eqref{eqn:theorem_easy_case_condition}. 
\end{corollary}
\begin{proof}
 When $f'(r(x;\theta))-\bar{\fnu}(r(x;\theta))\in\mathcal{L}[\bar{\eta}(x;\theta)]$ holds, 
 there exists a vector 
 $b\in\Real^d$ such that
 \[
 f'(r(x;\theta))-\bar{\fnu}(r(x;\theta))=b^T\bar{\eta}(x;\theta), 
 \]
 and thus    
\[
 \bar{c}^T U_{\bar{\eta}}^{-1}=\Enu[(f'(r(x;\theta))-\bar{\fnu}(r(x;\theta)))\nabla\log r(x;\theta)^T]
 \Enu[\bar{\eta}\nabla\log r^T]^{-1}=b^T
\]
 holds. Then we have 
 $\bar{c}^T U_{\bar{\eta}}^{-1}{\bar{\eta}}(x;\theta)=
 b^T{\bar{\eta}}(x;\theta)=f'(r(x;\theta))-\bar{\fnu}(r(x;\theta))$. 
  Hence
  \begin{align*}
   &\phantom{=}f(r(x;\theta))-(r(x;\theta)+\rho^{-1})(\bar{\fnu}(r(x;\theta))
   +\bar{c}^T U_{\bar{\eta}}^{-1}\bar{\eta}(x;\theta)) \\
   &=
   f(r(x;\theta))-(r(x;\theta)+\rho^{-1})f'(r(x;\theta))
   \in\mathcal{L}[\nabla\log r(x;\theta)]
  \end{align*}
 is satisfied, and thus \eqref{eqn:theorem_easy_case_condition} holds.  
\end{proof}

We consider the conjugate representation $f(r)=-f^*(f'(r))+rf'(r)$, that is, 
$\fde(r)=-f^*(f'(r))$ and $\fnu(r)=f'(r)$, 
where $f^*(r)=\sup_{s\in\Real}\,\{ rs-f(s) \}$. 
Then, Corollary \ref{corollary:other_optimal_condition} implies that the decomposition
based on the conjugate representation leads to an optimal estimator when 
the model $r(x;\theta)$ and the $f$-divergence satisfy
\begin{align}
  f(r(x;\theta))-(r(x;\theta)+\rho^{-1})f'(r(x;\theta))
 \in\mathcal{L}[\nabla\log{r}(x;\theta)]. 
 \label{eqn:conjugate-optimality-condition}
\end{align}
If \eqref{eqn:conjugate-optimality-condition} does not hold, the optimality of the
estimator based on the conjugate representation is not guaranteed. 
On the other hand, the decomposition \eqref{eqn:optimal-estimator-with-QinEstimator} leads
to an optimal estimator without specific conditions on the model and the $f$-divergence. 
In addition, when $\fnu(r)$ is defined as $\fnu(r)=f'(r)$, the moment matching estimator
using $\bar{\eta}(x;\theta)$ does not affect the asymptotic variance of the $f$-divergence
estimator. Indeed, the equality $f'(r(x;\theta))-\fnu(r(x;\theta))=0$ holds and the vector
$c$ in \eqref{eqn:f-div-asympto} vanishes. As a result, the variance of the estimator
$\widehat{D}_f$ depends only on the decomposition of $f$ up to the order $O_p(1)$. 

We show some examples for which Corollary \ref{corollary:QinEstimator_optimal}
and Corollary \ref{corollary:other_optimal_condition} are applicable to construct the
optimal estimator. 

\begin{example}[exponential density-ratio models and KL-divergence]
 \label{example:exp_model_KL-div}
 Let the model be $r(x;\theta)=\exp\{\theta^T\phi(x)\},\,\theta\in\Real^{d}$
 with $\phi(x)=(\phi_1(x),\ldots,\phi_{d}(x))^T$
 and $\phi_1(x)\equiv1$. 
 Then $\mathcal{L}[\nabla\log r(x;\theta)]$ is spanned by 
 $1,\phi_2(x),\ldots,\phi_{d}(x)$ and clearly $\mathcal{L}[\nabla\log r(x;\theta)]$
 includes the constant $1$.  
 The $f$-divergence with $f(r)=-\log r+r-1$ leads to KL-divergence. 
 Let $\fde(r)=-\log r-1$ and $\fnu(r)=1$, 
 then \eqref{eqn:cor_QinEstimator_Opt} is satisfied, since 
 \begin{align*}
  f(r(x;\theta))-(r(x;\theta)+\rho^{-1})\bar{\fnu}(r(x;\theta))
  =-\theta^T\phi(x)-1-\rho^{-1}\in\mathcal{L}[\nabla\log r(x;\theta)]
 \end{align*}
 holds. 
 Then, 
 we see that the function $\eta=\eta_{\mathrm{opt}}$ and the decomposition
 $\fde(r)=-\log r-1$ and $\fnu(r)=1$ lead to an optimal estimator of the KL-divergence. 
 We see that there is redundancy for the decomposition of $f$. 
 Indeed, for any constants $c_0, c_1\in\Real$, the function 
 $c_0+c_1\log r(x;\theta)$ is included in $\mathcal{L}[\nabla\log r]$. 
 Hence the decomposition 
 \begin{align*}
  \fnu(r)=\frac{r+c_1\log r+c_0}{r+\rho^{-1}},
  \qquad 
  \fde(r)=r-\log r-1-r\fnu(r)
 \end{align*}
 with $\bar{\eta}=\eta_{\mathrm{opt}}$ also leads to an optimal estimator. 
 The decomposition in \eqref{eqn:optimal-estimator-with-QinEstimator} is realized by
 setting $c_0=-1,\,c_1=-1$. 
\end{example}

\begin{example}[power-model and power-divergence]
 Let the model be
 $r(x;\theta)=\big(1+\alpha\theta^T \phi(x)\big)^{1/\alpha}$
 with $\phi_1(x)=1$, where $\alpha$ is the parameter to specify the divergence such that
 $\alpha>-1$. 
 Then $\mathcal{L}[\nabla\log{r}(x;\theta)]$ is the linear space spanned by
 $\phi_1(x)/r^\alpha,\ldots,\phi_d(x)/r^\alpha$. 
 We see that $1=(e_1+\alpha\theta)^T\nabla\log r(x;\theta)$ holds, where $e_1$ is
 the unit vector $(1,0,\ldots,0)^T\in\Real^d$. 
 The convex function $f(r)=r-1+(r^{-\alpha}-1)/\alpha$  leads to the power divergence 
 \cite{JRSS-B:Ali+Silvey:1966,SSM-Hungary:Csiszar:1967}, 
 \begin{align*}
  \int \pde(x)f\left(\frac{\pnu(x)}{\pde(x)}\right)dx=
  \frac{1}{\alpha}\bigg(\int\frac{\pde(x)^{\alpha+1}}{\pnu(x)^{\alpha}}dx-1\bigg). 
 \end{align*}
 Hellinger distance is given by setting $\alpha=-1/2$, and Pearson divergence is
 realized by setting $\alpha=1$. 
 In the limit of $\alpha\rightarrow0$, KL-divergence is recovered. 
 Letting $\fde(r)=-1+(r^{-\alpha}-1)/\alpha$ and $\fnu(r)=1$, we have
 \begin{align*}
  f(r(x;\theta))-(r(x;\theta)+\rho^{-1})\fnu(r(x;\theta))
  &=
  -\frac{\theta^T\phi(x)}{r^\alpha}-1-\rho^{-1}
  \in\mathcal{L}[\nabla\log{r}(x;\theta)]
 \end{align*}
 and thus, due to Corollary \ref{corollary:QinEstimator_optimal} 
 the decomposition 
 $\fde(r)=-1+(r^{-\alpha}-1)/\alpha,\,\fnu(r)=1$ and the moment matching estimator using 
 $\eta=\eta_{\mathrm{opt}}$ lead to an optimal estimator of the power divergence
 under the power model. Also, the decomposition
 \eqref{eqn:optimal-estimator-with-QinEstimator} leads to another optimal estimator. 
\end{example}

\begin{example}[exponential density-ratio model and mutual information]
\label{example:theorem2_exponential_density-ratio_model}
 Let the model be $r(x;\theta)=\exp\{\theta^T\phi(x)\},\,\theta\in\Real^{d}$
 with $\phi(x)=(\phi_1(x),\ldots,\phi_{d}(x))^T$
 and $\phi_1(x)\equiv1$. 
 Then, the linear space $\mathcal{L}[\nabla\log{r}(x;\theta)]$ is spanned by
 $\{\phi_1(x),\ldots,\phi_d(x)\}$
 and thus $\mathcal{L}[\nabla\log{r}(x;\theta)]$ includes the function of the form  
 $c_0+c_1\log r(x;\theta)$ for $c_0,c_1\in\Real$. 
 Let the convex function $f(r)$ be
 \begin{align}
  f(r)&=\frac{1}{1+\rho}\log\frac{1+\rho}{1+\rho r}
  +r\,\frac{\rho}{1+\rho}\log\frac{r(1+\rho)}{1+\rho r}
  \label{eqn:example-fdiv-mutual-info}
 \end{align}
 for $\rho>0$. 
 Then the corresponding $f$-divergence is reduced to mutual information: 
 \begin{align*}
 \int \pde(x)f\left(\frac{\pnu(x)}{\pde(x)}\right)dx
  =
  \int\!\sum_{y=\mathrm{n},\mathrm{d}} p(x,y)\log\frac{p(x,y)}{p(x)p(y)}dx,
 \end{align*}
 in which the joint probability is defined as
 \begin{align*}
  p(x,\mathrm{n})=\pnu(x)\frac{\rho}{1+\rho},\quad 
  p(x,\mathrm{d})=\pde(x)\frac{1}{1+\rho}. 
 \end{align*}
 The equality $\pde=\pnu$ implies that the conditional probability $p(x|y)$ is independent
 of $y$. Thus, mutual information becomes zero if and only if $\pde=\pnu$ holds. 
 For any moment matching estimator, the following decomposition satisfies 
 the condition in Corollary \ref{corollary:other_optimal_condition}: 
 \begin{align}
  \fde(r)=\frac{1}{1+\rho}\log\frac{1+\rho}{1+\rho r},\qquad 
  \fnu(r)=\frac{\rho}{1+\rho}\log\frac{r(1+\rho)}{1+\rho r}. 
  \label{eqn:mutual-info-decomp}
 \end{align}
 Indeed, the equalities
 \begin{align*}
  &
  f(r(x;\theta))-(r(x;\theta)+\rho^{-1})f'(r(x;\theta))=\frac{-\log(r(x;\theta))}{1+\rho}
  \in\mathcal{L}[\nabla\log{r}(x;\theta)],\\
  & f'(r(x;\theta))-\fnu(r(x;\theta))=0\in\mathcal{L}[\eta(x;\theta)]
 \end{align*}
 hold for any $\eta(x;\theta)$. 
 The estimator derived from the decomposition above with $\eta=\eta_{\mathrm{opt}}$ has
 also been proposed by Keziou and Leoni-Aubin \cite{kezioua05:_test_of_homog_in_semip}. In
 their work, the estimator is derived as the conjugate expression of 
 the prospective likelihood. 
 In this example, we present another
 characterization, that is, the optimal estimator for mutual information. 
\end{example}

\begin{example}[linear model]
\label{example:theorem2_linear-model}
 Let $r(x;\theta)=1+\theta^T\phi(x)$ and $\phi_1(x)\equiv1$. The subspace
 $\mathcal{L}[\nabla\log r(x;\theta)]$ is spanned by $\{\phi_1/r,\ldots,\phi_d/r\}$, and
 thus $\mathcal{L}[\nabla\log r(x;\theta)]$ includes the function of the form $c_0+c_1/r$
 for $c_0,c_1\in\Real$. Let the convex function $f$ be 
 \begin{align*}
  f(r)=\frac{1}{\rho+1}\bigg[r-1+(1+\rho r)\log\frac{1+\rho r}{r(1+\rho)}\bigg]
 \end{align*}
 for $\rho>0$. 
 Then the corresponding $f$-divergence is expressed as
\begin{align*}
 \int \pde(x)f\left(\frac{\pnu(x)}{\pde(x)}\right)dx
 =\mathrm{KL}\left(\frac{\pde+\rho\,\pnu}{1+\rho},\,\pnu\right), 
\end{align*}
 where $\mathrm{KL}$ is the Kullback-Leibler divergence. 
 The $f$-divergence vanishes if and only if $\pnu=\pde$ holds. 
 Using Corollary \ref{corollary:QinEstimator_optimal}, we see that the decomposition 
 \begin{align*}
  \fde(r)=
  \frac{1}{\rho+1}\bigg[\frac{r-1}{1+\rho r}+\log\frac{1+\rho r}{r(1+\rho)}\bigg],
  \qquad
  \fnu(r)=
  \frac{\rho}{\rho+1}\bigg[
  \frac{r-1}{1+\rho r} +\log\frac{1+\rho r}{r(1+\rho)}\bigg],
 \end{align*}
 and the moment matching estimator using $\eta=\eta_{\mathrm{opt}}$ lead to an optimal
 estimator for the above $f$-divergence.  
 On the other hand, due to Corollary \ref{corollary:other_optimal_condition},
 we see that the decomposition 
 \begin{align*}
  \fde(r)=\frac{1}{1+\rho}\log\frac{1+\rho r}{r(1+\rho)},\quad 
  \fnu(r)=f'(r)=\frac{1}{r(1+\rho)}\bigg[r-1+\rho r\log\frac{1+\rho r}{r(1+\rho)}\bigg]
 \end{align*}
 leads to another optimal estimator. 
\end{example}

\section{Homogeneity test exploiting $f$-divergence estimator}
\label{sec:Higher-order_Asymptotics} 
For the homogeneity test of $\pnu$ and $\pde$, we need to know the asymptotic distribution 
of $\widehat{D}_f$ under the null hypothesis of \eqref{eqn:homogeneity_test}. 
In this section, we assume 
\begin{align*}
 \frac{\pnu(x)}{\pde(x)}=r(x;\theta^*)\equiv1
\end{align*}
and $1\in\mathcal{L}[\eta(x;\theta^*)]$. 
Then we see that $\Pbb{f}=0$ holds for any decomposition of $f$, 
and thus the asymptotic expansion of $\widehat{D}_f$ around $\theta=\theta^*$ satisfies 
\begin{align*}
 \sqrt{m}(\widehat{D}_f-D_f) =o_p(1),
\end{align*}
where $D_f=D_f(\pde,\pnu)=0$. 
For $\pnu=\pde$, the variance covariance matrix $(\rho\Vnu[\eta]+\Vnu[\eta])/(\rho+1)$ 
in \eqref{eqn:asymptotics-parameter} is degenerate. 
This is the reason why the probabilistic order of $\sqrt{m}(\widehat{D}_f-D_f)$ becomes
$o_p(1)$. On the other hand, for $\pnu\neq\pde$, 
$\sqrt{\mde}(\widehat{D}_f-D_f)$ is of the order $O_p(1)$. 

Below, we consider the optimal estimator $\widehat{D}_f$ defined from 
\eqref{eqn:optimal-estimator-with-QinEstimator}. The asymptotic distribution of the
optimal estimator is given by the following theorem. 
\begin{theorem}
 \label{theorem:min-variance-higher-order-expansion}
 Let Assumption \ref{cond:model-assumption} hold, and we assume
 $\pnu(x)/\pde(x)=r(x;\theta^*)=1$. 
 Suppose that the ratio of the sample size, $\rho=\mnu/\mde$, converges to a positive
 value, and that the $d$ by $d$ symmetric matrix $U_\eta$ with $\eta=\eta_{\mathrm{opt}}$
 is non-degenerate in the vicinity of $\theta=\theta^*$. Let $\widehat{D}_f$ be the
 estimator defined from \eqref{eqn:optimal-estimator-with-QinEstimator}. 
 Then, in terms of the asymptotic distribution of $\widehat{D}_f$, we obtain 
 \begin{align*}
  \frac{2m}{f''(1)}
  \widehat{D}_f  \ \cd\  \chi_{d-1}^2, 
 \end{align*}
 where $\chi_{\ell}^2$ is the chi-square distribution with $\ell$ degrees of freedom. 
\end{theorem}
The proof is deferred to Appendix 1. 
For the homogeneity test of $\pnu$ and $\pde$, 
the null hypothesis $\pnu=\pde$ is rejected if  
\begin{align}
 \widehat{D}_f  \geq 
 \frac{f''(1)}{2m}
 \chi_{d-1}^2(1-\alpha)
 \label{eqn:rejection-condition}
\end{align}
is satisfied, where $\chi_{d-1}^2(1-\alpha)$ is the chi-square $100(1-\alpha)$ percent
point function with  $d-1$ degrees of freedom. 
The homogeneity test based on \eqref{eqn:rejection-condition} with the optimal 
choice \eqref{eqn:optimal-estimator-with-QinEstimator} is referred to as 
\emph{$\widehat{D}_f$-based test}. 

We consider the power function of the homogeneity test, and compare the proposed method to
the other method. A standard approach for the homogeneity test is exploiting the
asymptotic distribution of the empirical likelihood estimator $\widehat{\theta}$. 
Under the model 
\begin{align}
r(x;\theta)=\exp\{\alpha+\phi(x;\beta)\},\qquad
 \theta=(\alpha,\beta)\in\Real\times \Real^{d-1}, 
 \label{eqn:ratio_exp_model}
\end{align}
Fokianos et al.~\cite{fokianos01:_semip_approac_to_one_way_layout} pointed out that
the asymptotic distribution of the empirical likelihood estimator 
$\widehat{\theta}=(\widehat{\alpha},\,\widehat{\beta})\in\Real\times\Real^{d-1}$ 
under the null hypothesis $\pnu=\pde$ is given as 
\begin{align*}
 \sqrt{m}(\widehat{\beta}-\beta^*)\ \cd\ 
 N_{d-1}(\,0,\Vnu[\nabla_{\beta}\phi]^{-1}\,),
\end{align*}
where $\theta^*=(\alpha^*,\beta^*)$ and $\nabla_\beta\phi$ is the $d-1$ dimensional
gradient vector of $\phi(x;\beta)$ at $\beta=\beta^*$ with respect to the parameter
$\beta$. 
Then the null hypothesis is rejected if the test statistic 
\begin{align}
 S=m (\widehat{\beta}-\beta^*)^T \Vnuh[\nabla_{\beta}\phi](\widehat{\beta}-\beta^*)
 \label{eqn:test-stat-Fokianos}
\end{align}
is larger than $\chi_{d-1}^2(1-\alpha)$, where $\Vnuh[\nabla_{\beta}\phi]$ is a consistent
estimator of $\Vnu[\nabla_{\beta}\phi]$. In this paper, the homogeneity test based on the
statistic $S$ is referred to as \emph{empirical likelihood test}. 
Fokianos et al.~\cite{fokianos01:_semip_approac_to_one_way_layout} studied
statistical properties of empirical likelihood test through numerical experiments,
and reported that the power of empirical likelihood test is comparable to standard
$t$-test and $F$-test. 

Below, we show that the power of $\widehat{D}_f$-based test is the same as empirical
likelihood test under the setup of local alternative,  
where the distributions $\pnu$ and $\pde$ vary according to the sample size. 
To compute the power function, we assume the following conditions. 
\begin{assumption}
 \label{assumption:local_alternative}
 Let the density ratio model $r(x;\theta)$ be represented as \eqref{eqn:ratio_exp_model}. 
 Let $r(x;\theta^*)=1$ and $\theta_{m}=\theta^*+h_{m}/\sqrt{m}$, 
 where $h_m\in\Real^d$ and $\lim_{m\rightarrow\infty}h_{m}=h\in\Real^d$.
 Suppose $\pde(x)=p(x)$ for a fixed probability density $p(x)$ 
 and that the probability density $\pnu^{(m)}$ is represented as 
 $\pnu^{(m)}(x)=\pde(x)r(x;\theta_m)$.  
 For each sample size $\mnu$ and $\mde$, 
 the samples $\xnu_1,\ldots,\xnu_{\mnu}$ are generated from 
 $\pnu^{(m)}$, and 
 $\xde_1,\ldots,\xde_{\mde}$ are generated from $\pde$. 
 The limit of the ratio $\mnu/\mde$ is denoted as $\rho$. 
 Let the matrix-valued function $M(\theta)$ and $U(\theta)$ be
 \begin{align*}
  M(\theta)&=\int p(x) \nabla\log r(x;\theta)\nabla\log r(x;\theta)^T dx,\\
  U(\theta)&=\int p(x)
  \frac{1}{1+\rho r(x;\theta)}\nabla\log r(x;\theta)\nabla\log r(x;\theta)^Tdx, 
 \end{align*}
 and assume that these are continuous and non-degenerate in the vicinity of $\theta^*$. 
 Let $V[\nabla r]$ be the variance-covariance matrix of
 $\nabla\log{r}(x;\theta^*)=\nabla{r}(x;\theta^*)$ under $p(x)$. We assume 
 \begin{align}
  \frac{1}{\mnu}\sum_{j=1}^{\mnu}\nabla{r}(\xnu_j;\theta^*)\nabla{r}(\xnu_j;\theta^*)^T
  &\ \ \cp\ \ M(\theta^*),
  \label{eqn:local_alternative_expectation}\\
  \sqrt{m}\,U(\theta_{m})(\widehat{\theta}-\theta_{m})
  &\ \stackrel{\theta_{m}}{\longrightarrow}\ 
  N\bigg(0,\frac{1}{(1+\rho)^2}V[\nabla r]\bigg), 
  \label{eqn:local_alternative_CLT}
 \end{align}
 where  \eqref{eqn:local_alternative_expectation} implies the convergence in probability,
 that is, for any $\varepsilon>0$, the probability such that
\begin{align*}
 \bigg|\frac{1}{\mnu}\sum_{j=1}^{\mnu}\nabla{r}(\xnu_j;\theta^*)\nabla{r}(\xnu_j;\theta^*)^T
 -M(\theta^*)  \bigg|>\varepsilon
 \end{align*}
 under the samples from $\pde^{(m)}$ converges to zero when $\mnu$ tends to infinity. 
 The notation $X_m\stackrel{\theta_m}{\longrightarrow} P$ in \eqref{eqn:local_alternative_CLT} 
 denotes that the distribution function of $X_{m}$ depending on $\pnu^{(m)}$ and $\pde$ converges to
 $P$ in law, when $m$ tends to infinity. 
 See Section~14 in \cite{van98:_asymp_statis} and Section~11.4.2 in
 \cite{lehmann05:_testin_statis_hypot} for details of the asymptotic theory under the
 local alternative. 
 For $h_{m}=0\in\Real^d$, the condition on 
 $\sqrt{m}U(\theta_{m})(\widehat{\theta}-\theta_{m})$ is reduced to
 \eqref{eqn:asymptotics-parameter} with $\eta=\eta_{\mathrm{opt}}$. 
\end{assumption}

 In the above, one can make the assumption weaker such that the probability $\pde$ also
 varies according to the sample size. 
 We adopt the simplified assumption above to avoid technical difficulties. 
 \begin{theorem}
  \label{theorem:power-function}
  Under 
  Assumption \ref{cond:model-assumption} and 
  Assumption \ref{assumption:local_alternative}, 
  the power function of $\widehat{D}_f$-based test is
  asymptotically given as $\Pr\big\{ Y \geq \chi^2_{d-1}(1-\alpha)\big\}$, where $Y$ is 
  the random variable whose distribution function is the non-central chi-square
  distribution with $d-1$ degrees of freedom and non-centrality parameter 
  $h^T M(\theta^*)h$. 
  Moreover, the asymptotic power function of empirical likelihood test
  is the same. 
 \end{theorem}
 The proof is given in Appendix 2. 
 Theorem \ref{theorem:power-function} implies that, under the local alternative, the power
 function of $\widehat{D}_f$-based test does not depend on choice of the
 $f$-divergence and that empirical likelihood test has the same power as
 $\widehat{D}_f$-based test. 

 Next, we consider the power function under the misspecification case. 
\begin{theorem}
 \label{theorem:power_misspecifiedmodel}
 We assume that the density ratio $\pnu^{(m)}/\pde$ is not realized by the model $r(x;\theta)$, 
 and that $\pnu^{(m)}$ is represented as 
  \begin{align*}
   \pnu^{(m)}(x)=\pde(x)
   \bigg(r(x;\theta_m)+\frac{s_m(x)+\varepsilon_m}{\sqrt{m}}\bigg), 
  \end{align*}
 where $s_m(x)$ satisfies $E[s_m(x)]=0$ under the probability $\pde(x)=p(x)$, and 
 assume $\lim_{m\rightarrow\infty}\varepsilon_m=\varepsilon$. 
 Suppose Assumption \ref{cond:model-assumption} and 
 Assumption \ref{assumption:local_alternative} except the definition of $\pnu^{(m)}(x)$. 
 Then, under the setup of the local alternative,
 the power function of $\widehat{D}_f$-based test is larger than or equal to that of
 empirical likelihood test. 
\end{theorem}
 The proof is given in Appendix 3. Even in the misspecification case, 
 the assumption \eqref{eqn:local_alternative_expectation} and
 \eqref{eqn:local_alternative_CLT} will be valid, since eventually the limit of
 $\pnu^{(m)}/\pde$ is realized by the model $r(x;\theta^*)=1$. 
 Theorem \ref{theorem:power-function} and Theorem  \ref{theorem:power_misspecifiedmodel}
 indicate that $\widehat{D}_f$-based test is more powerful than empirical
 likelihood test regardless of whether the model $r(x;\theta)$ is correct or slightly
 misspecified.

\section{Numerical Studies}
\label{sec:Numerical_Studies}
In this section, we report numerical results for  illustrating the adequacy of the asymptotic
theory for finite-sample inference. 

We examine two $f$-divergences for the homogeneity test. 
One is KL-divergence defined by $f(r)=r-1-\log(r)$ as shown in Example
\ref{example:exp_model_KL-div}, and the test statistic is derived from 
\eqref{eqn:optimal-estimator-with-QinEstimator}. 
This is referred to as \emph{KL-based test}. 
The other is mutual information defined by \eqref{eqn:example-fdiv-mutual-info}, and
the estimator $\widehat{D}_f$ is derived from the decomposition
\eqref{eqn:mutual-info-decomp} and the moment matching estimator
$\eta=\eta_{\mathrm{opt}}$. This is referred to as \emph{MI-based test}. 
These tests are compared to empirical likelihood test \eqref{eqn:test-stat-Fokianos}
proposed by 
Fokianos et al.~\cite{fokianos01:_semip_approac_to_one_way_layout} and Hotelling
$T^2$-test. 
The null hypothesis of the testing is $H_0:\pnu=\pde$ and the alternative is
$H_1:\pnu\neq\pde$. 
The type-I error and the power function of these tests are computed. 
In all numerical studies, the sample $x$ is $10$-dimensional vector, and 
the semiparametric model for density ratio is defined as 
\begin{align}
 r(x;\theta)=
 \exp\bigg\{
 \alpha+\sum_{i=1}^{10}\beta_i x_i+\sum_{j=1}^{10}\beta_{10+j}\, x_j^2
 \bigg\}
 \label{eqn:sim-ratio-model}
\end{align}
with the $21$-dimensional parameter $\theta=(\alpha,\beta_1,\ldots,\beta_{20})$. 

First we assume that the null hypothesis $\pnu=\pde$ is correct, and we compute the
type-I error. 
We consider three cases: in the first case, the distributions of $\pnu$ and $\pde$ are
given as the $10$-dimensional normal distribution $N_{10}(0,I_{10})$; in the second case,
each element of $x\in\Real^{10}$ is independent and identically distributed from the
$t$-distribution with $10$ degrees of freedom; and in the third case, each element of
$x\in\Real^{10}$ is independent and identically distributed from the $t$-distribution
with $ 5$ degrees of freedom. The sample size is set to $\mnu=\mde$ and varies 
from 100 to 1200, and the significance level is set to $0.05$. 
The type-I errors are averaged over $300$ runs. 
For each case, the averaged type-I errors of KL-based test, MI-based test, and 
empirical likelihood test are shown in 
Table~\ref{table:typeI_error_plot}. 
In the normal case, the type-I error of three tests converges to the significance level
with modest sample size. 
In the case of $t$-distribution, the type-I error of empirical likelihood test is
larger than the significance level even with large sample size. 
On the other hand the type-I error of MI-based test is close to the significance level
with moderate sample size even for the case of $t$-distribution. 

\begin{table}[tb]
 \caption{Averaged Type-I errors over 300 runs are shown as functions of the number of samples. 
 Normal distribution, $t$-distribution with 10 degrees of freedom, and $t$-distribution with 5 degrees of freedom 
 are examined as $\pnu$ and $\pde$. 
 Below, ``MI'', ``KL'' and ``emp.'' denote MI-based test, KL-based test and empirical likelihood test, respectively. }
 \label{table:typeI_error_plot}
 \centering\vspace*{4mm}
 \hspace*{-14mm}
 \begin{tabular}{lrrrcrrrcrrr}
     & \multicolumn{3}{c}{$10$-dim Normal} 
  & & \multicolumn{3}{c}{$10$-dim. $t$-dist. (df=$10$)} 
  & & \multicolumn{3}{c}{$10$-dim. $t$-dist. (df=$5$)} \\ \cline{2-4}\cline{6-8}\cline{10-12}
 $\mnu(=\mde)$&     MI &      KL &      emp. &&       MI     &  KL &     emp. &&      MI &    KL &  emp.\\
100  &           0.080 &   0.117 &     0.183  &&  0.133  &  0.217  & 0.297    &&   0.100 & 0.210 & 0.377\\
500  &           0.070 &   0.083 &     0.080  &&  0.070  &  0.090  & 0.107    &&   0.060 & 0.107 & 0.187\\
1000 &           0.053 &   0.057 &     0.060  &&  0.073  &  0.070  & 0.093    &&   0.070 & 0.103 & 0.170\\
1200 &           0.047 &   0.050 &     0.067  &&  0.073  &  0.087  & 0.097    &&   0.067 & 0.093 & 0.170\\
  \hline
  \end{tabular}
\vspace*{4mm}
\end{table}

Next, we compute the power function of KL-based test, MI-based test, 
empirical likelihood test, and Hotelling $T^2$-test. In the numerical simulations,
$\pnu(x)$ is fixed and $\pde(x)$ is varied by changing the mean parameter or
the scale parameter. 
In the same way as the computation of the type-I error, 
$\pnu(x)$ is fixed to one of the three probabilities: $10$-dimensional normal distribution
$N_{10}(0,I_{10})$, $10$-dimensional $t$-distortion with 10 or 5 degrees of freedom. 
The probability $\pde(x)$ is defined by changing the mean or the variance of the
probability $\pnu(x)$. 
In the first setup, the sample $\xde=(\xde_1,\ldots,\xde_{10})$ from $\pde$ is
computed such that  
\begin{align}
 \xde_\ell=x_\ell+\mu,\ \ \ell=1,\ldots,10,\qquad
 x=(x_1,\ldots,x_{10})\sim \pnu, 
 \label{eqn:mean-shift}
\end{align}
that is, the mean parameter $\mu\in\Real$ is added to each element of $x$. 
Hence, $\pnu=\pde$ holds for $\mu=0$. 
In the second setup, the sample $\xde=(\xde_1,\ldots,\xde_{10})$ from $\pde$ is
computed such that 
\begin{align}
 \xde_\ell=\sigma\times x_\ell,\ \ \ell=1,\ldots,10,\qquad
 x=(x_1,\ldots,x_{10})\sim \pnu, 
 \label{eqn:scale-shift}
\end{align}
that is, the scale parameter $\sigma>0$ is multiplied to each element of $x$. 
Hence, $\pnu=\pde$ holds for $\sigma=1$. 
In all simulations, the sample size is set to $\mnu=\mde=500$ or $1000$ and 
the significance level is $0.05$. 
When both $\pnu$ and $\pde$ are the multi-dimensional normal distribution, the density
ratio model \eqref{eqn:sim-ratio-model} includes the true ratio. For the $t$-distribution,
however, the true ratio $r(x)$ resides outside of the model \eqref{eqn:sim-ratio-model}. 
The power functions are averaged over $300$ runs. 

Table~\ref{table:typeII_error_mu} shows the averaged power functions over $300$ runs 
for the setup \eqref{eqn:mean-shift}. 
The mean parameter $\mu$ varies from $-0.1$ to $0.1$. When both $\pnu$ and $\pde$ are the
normal distribution, the power functions of KL-based test, MI-based test, and
empirical likelihood almost coincide with each other.
The power of Hotelling $T^2$-test is slightly larger 
than the others. This result is obvious, since Hotelling $T^2$-test works well under
the normal distribution. 
Under the $t$-distribution with $5$ degree of freedom, 
the power of empirical likelihood test around $\mu=0$ is much larger than the
significance level.
That is, empirical likelihood test is not conservative, and will lead false
positive with high probability. 
In MI-based test, the power around $\mu=0$ is close to the significance level and 
the power is comparable to Hotelling $T^2$-test outside of the vicinity of $\mu=0$. 

Table~\ref{table:typeII_error_sigma} shows the averaged power functions over $300$ runs 
when the scale parameter $\sigma$ in \eqref{eqn:scale-shift} varies from $0.9$ to $1.1$. 
In this case, the means of $\pnu$ and $\pde$ are the same, and hence Hotelling $T^2$-test 
fails to detect the difference of $\pnu$ and $\pde$. 
In addition, we see that the power function of empirical likelihood test is biased, 
that is, the power function takes the minimum value at $\sigma$ less than $1$. 
This is because the estimated variance, $\Vnuh$, based on empirical likelihood
estimator tends to take slightly small values than the true variance. 
In MI-based test, the power around $\sigma=1$ is close to the significance level,
while the power of KL-based test is slightly larger than the significance level 
around $\sigma=1$. 

As shown above, when the model $r(x;\theta)$ is correct, the power of KL-based test,
MI-based test, and empirical likelihood test is almost the same. 
Thus, the numerical simulations meet the
theoretical results in Theorem \ref{theorem:power-function}. 
Empirical likelihood test has large type-I error and the power is slightly biased
especially when the samples are generated from the $t$-distribution. 
Throughout the simulations, MI-based test has the comparable power to the other
methods, while the type-I error is well controlled. 
In the simulations, we see that the null distribution of MI-based test is 
approximated by the asymptotic distribution more accurately than that of KL-based test, 
although the first-order asymptotic theory provided in Section~\ref{sec:Higher-order_Asymptotics}
does not explain the difference between MI-based test and KL-based test. 
We expect that higher order asymptotic theory is needed to
better understand the difference among $f$-divergences for the homogeneity test. 

\begin{table}[tb]
 \caption{
 Averaged power functions over $300$ runs are shown as functions of the mean parameter of
 the probability 
 $\pde(x)$, where $\pde(x)$ is defined by \eqref{eqn:mean-shift} through the probability
 $\pnu$.  
 Normal distribution, $t$-distribution with 10 degrees of freedom, and $t$-distribution with 5 degrees of freedom 
 are examined as $\pnu$. 
 Below, ``MI'', ``KL'', ``emp.'' and ``Hote.'' denote MI-based test, KL-based test,
 empirical likelihood test and Hotelling $T^2$-test, respectively. }
 \label{table:typeII_error_mu}
 \centering
 \footnotesize\vspace*{3mm}\hspace*{-15mm}
 \begin{tabular}{crrrrcrrrrcrrrr}
  \multicolumn{15}{c}{$\mnu=\mde=500$}\\ \hline
    & \multicolumn{4}{c}{$10$-dim Normal} 
  & & \multicolumn{4}{c}{$10$-dim. $t$-dist. (df=$10$)} 
  & & \multicolumn{4}{c}{$10$-dim. $t$-dist. (df=$5$)} \\ \cline{2-5}\cline{7-10}\cline{12-15}
\multicolumn{1}{c}{ $\mu$} &   MI  &    KL &    emp. &  Hote.  &&    MI   &   KL &    emp. &  Hote.  &&    MI &   KL &    emp. &  Hote. \\
-0.1   & 0.894 & 0.902 &   0.898 &  0.964  &&   0.812 &0.826 &   0.822 &  0.886  && 0.680 &0.724 &   0.746 &  0.750\\
-0.08  & 0.650 & 0.662 &   0.654 &  0.778  &&	0.538 &0.572 &   0.608 &  0.714  && 0.472 &0.532 &   0.592 &  0.562\\
-0.06  & 0.362 & 0.388 &   0.384 &  0.510  &&	0.302 &0.328 &   0.360 &  0.418  && 0.236 &0.296 &   0.408 &  0.258\\
-0.04  & 0.184 & 0.190 &   0.214 &  0.226  &&	0.132 &0.156 &   0.200 &  0.176  && 0.130 &0.186 &   0.284 &  0.134\\
-0.02  & 0.084 & 0.100 &   0.104 &  0.074  &&	0.080 &0.104 &   0.148 &  0.080  && 0.082 &0.132 &   0.216 &  0.072\\
 0     & 0.046 & 0.058 &   0.062 &  0.036  &&	0.062 &0.082 &   0.098 &  0.046  && 0.054 &0.090 &   0.170 &  0.056\\
 0.02  & 0.072 & 0.080 &   0.092 &  0.070  &&	0.064 &0.076 &   0.104 &  0.044  && 0.090 &0.150 &   0.218 &  0.074\\
 0.04  & 0.196 & 0.206 &   0.212 &  0.210  &&	0.138 &0.160 &   0.186 &  0.158  && 0.138 &0.200 &   0.304 &  0.130\\
 0.06  & 0.374 & 0.398 &   0.424 &  0.490  &&	0.314 &0.348 &   0.372 &  0.388  && 0.260 &0.332 &   0.380 &  0.274\\
 0.08  & 0.658 & 0.688 &   0.698 &  0.760  &&	0.528 &0.554 &   0.586 &  0.632  && 0.470 &0.536 &   0.578 &  0.528\\
 0.1   & 0.866 & 0.878 &   0.870 &  0.954  &&	0.796 &0.810 &   0.814 &  0.878  && 0.672 &0.740 &   0.750 &  0.760\\ \hline\\
  \multicolumn{15}{c}{$\mnu=\mde=1000$}\\ \hline
    & \multicolumn{4}{c}{$10$-dim Normal} 
  & & \multicolumn{4}{c}{$10$-dim. $t$-dist. (df=$10$)} 
  & & \multicolumn{4}{c}{$10$-dim. $t$-dist. (df=$5$)} \\ \cline{2-5}\cline{7-10}\cline{12-15}
\multicolumn{1}{c}{ $\mu$} &  
         MI  &    KL &     emp. &  Hote. &&    MI &   KL &    emp. &  Hote.  &&    MI  &   KL &    emp. &  Hote. \\
-0.1   & 0.996& 0.998&    0.998 & 1.000  &&  0.996& 0.996&    0.998&  0.994  &&   0.958& 0.964&   0.968 &  0.990 \\
-0.08  & 0.952& 0.954&    0.954 & 0.986	 &&  0.902& 0.906&    0.906&  0.960  &&	  0.790& 0.816&   0.824 &  0.864 \\
-0.06  & 0.694& 0.698&    0.698 & 0.794	 &&  0.616& 0.634&    0.652&  0.784  &&	  0.470& 0.516&   0.550 &  0.594 \\
-0.04  & 0.320& 0.336&    0.336 & 0.422	 &&  0.258& 0.278&    0.304&  0.340  &&	  0.208& 0.246&   0.316 &  0.232 \\
-0.02  & 0.096& 0.110&    0.122 & 0.132	 &&  0.080& 0.090&    0.102&  0.110  &&	  0.094& 0.128&   0.220 &  0.100 \\
 0     & 0.058& 0.060&    0.064 & 0.044	 &&  0.058& 0.068&    0.102&  0.052  &&	  0.074& 0.100&   0.166 &  0.068 \\
 0.02  & 0.088& 0.090&    0.098 & 0.100	 &&  0.112& 0.120&    0.142&  0.114  &&	  0.092& 0.128&   0.194 &  0.078 \\
 0.04  & 0.308& 0.322&    0.324 & 0.472	 &&  0.296& 0.318&    0.324&  0.396  &&	  0.222& 0.258&   0.314 &  0.248 \\
 0.06  & 0.724& 0.730&    0.728 & 0.836	 &&  0.622& 0.640&    0.652&  0.752  &&	  0.474& 0.500&   0.538 &  0.586 \\
 0.08  & 0.956& 0.960&    0.958 & 0.978	 &&  0.890& 0.900&    0.904&  0.962  &&	  0.770& 0.790&   0.818 &  0.856 \\
 0.1   & 0.998& 0.996&    0.998 & 1.000	 &&  0.992& 0.990&    0.988&  0.998  &&   0.966& 0.970&   0.980 &  0.988 \\\hline
  \end{tabular}
\end{table}

\begin{table}[tb]
 \caption{
 Averaged power functions over $300$ runs are shown as functions of the scale parameter of the probability $\pde(x)$, 
 where $\pde(x)$ is defined by \eqref{eqn:scale-shift} through the probability $\pnu$. 
 Normal distribution, $t$-distribution with 10 degrees of freedom, and $t$-distribution with 5 degrees of freedom 
 are examined as $\pnu$. 
 Below, ``MI'', ``KL'', ``emp.'' and ``Hote.'' denote MI-based test, KL-based test,
 empirical likelihood test and Hotelling $T^2$-test, respectively. }
 \label{table:typeII_error_sigma}
 \centering
  \footnotesize\vspace*{3mm}\hspace*{-15mm}
 \begin{tabular}{crrrrcrrrrcrrrr}
  \multicolumn{15}{c}{$\mnu=\mde=500$}\\ \hline
    & \multicolumn{4}{c}{$10$-dim Normal} 
  & & \multicolumn{4}{c}{$10$-dim. $t$-dist. (df=$10$)} 
  & & \multicolumn{4}{c}{$10$-dim. $t$-dist. (df=$5$)} \\ \cline{2-5}\cline{7-10}\cline{12-15}
\multicolumn{1}{c}{ $\sigma$} & 
           MI &    KL &    emp. &  Hote.  &&    MI   &   KL &    emp. &  Hote.  &&    MI &   KL &   emp. &  Hote. \\
0.9  &  1.000&  0.998&    0.994 &  0.042  &&  0.986& 0.978 &   0.912  &   0.070 &&0.846 &0.788  &  0.484 &  0.044 \\
0.92 &  0.976&  0.976&    0.948 &  0.046  &&  0.850& 0.786 &   0.638  &   0.058	&&0.592 &0.492  &  0.204 &  0.036 \\
0.94 &  0.750&  0.714&    0.554 &  0.044  &&  0.552& 0.486 &   0.282  &   0.034	&&0.328 &0.272  &  0.110 &  0.042 \\
0.96 &  0.354&  0.328&    0.184 &  0.054  &&  0.240& 0.186 &   0.096  &   0.048	&&0.208 &0.172  &  0.094 &  0.054 \\
0.98 &  0.112&  0.096&    0.054 &  0.050  &&  0.078& 0.070 &   0.054  &   0.042	&&0.084 &0.114  &  0.142 &  0.060 \\
1    &  0.064&  0.078&    0.080 &  0.048  &&  0.052& 0.074 &   0.098  &   0.042	&&0.066 &0.086  &  0.170 &  0.030 \\
1.02 &  0.102&  0.146&    0.212 &  0.044  &&  0.104& 0.164 &   0.248  &   0.056	&&0.090 &0.182  &  0.338 &  0.054 \\
1.04 &  0.334&  0.406&    0.516 &  0.046  &&  0.218& 0.316 &   0.490  &   0.050	&&0.158 &0.314  &  0.514 &  0.050 \\
1.06 &  0.666&  0.744&    0.840 &  0.050  &&  0.516& 0.670 &   0.818  &   0.060	&&0.324 &0.528  &  0.780 &  0.054 \\
1.08 &  0.946&  0.976&    0.992 &  0.044  &&  0.806& 0.876 &   0.948  &   0.038	&&0.538 &0.716  &  0.862 &  0.060 \\
1.1  &  0.992&  0.994&    0.998 &  0.064  &&  0.966& 0.992 &   0.998  &   0.032	&&0.774 &0.868  &  0.974 &  0.046 \\\hline\\
  \multicolumn{15}{c}{$\mnu=\mde=1000$}\\ \hline
    & \multicolumn{4}{c}{$10$-dim Normal} 
  & & \multicolumn{4}{c}{$10$-dim. $t$-dist. (df=$10$)} 
  & & \multicolumn{4}{c}{$10$-dim. $t$-dist. (df=$5$)} \\ \cline{2-5}\cline{7-10}\cline{12-15}
\multicolumn{1}{c}{ $\sigma$} &  
         MI  &    KL &   emp. &  Hote. &&    MI &   KL &    emp. &  Hote.  &&    MI  &   KL &    emp. &  Hote. \\
0.9  & 1.000 &1.000  &  1.000 &  0.062 &&   1.000& 1.000&    1.000&   0.046&&   0.992& 0.986&    0.914&  0.056 \\
0.92 & 1.000 &1.000  &  1.000 &  0.074 &&   0.998& 0.996&    0.984&   0.052&&   0.892& 0.854&    0.638&  0.054 \\
0.94 & 0.982 &0.980  &  0.968 &  0.054 &&   0.912& 0.892&    0.766&   0.074&&	0.620& 0.532&    0.294&  0.054 \\
0.96 & 0.648 &0.608  &  0.502 &  0.052 &&   0.464& 0.412&    0.278&   0.040&&	0.264& 0.214&    0.118&  0.058 \\
0.98 & 0.148 &0.132  &  0.104 &  0.042 &&   0.118& 0.104&    0.072&   0.054&&	0.108& 0.098&    0.080&  0.058 \\
1    & 0.046 &0.050  &  0.058 &  0.030 &&   0.054& 0.060&    0.074&   0.040&&	0.066& 0.088&    0.164&  0.048 \\
1.02 & 0.170 &0.200  &  0.256 &  0.058 &&   0.120& 0.158&    0.256&   0.040&&	0.096& 0.158&    0.310&  0.046 \\
1.04 & 0.678 &0.732  &  0.806 &  0.060 &&   0.416& 0.532&    0.650&   0.070&&	0.272& 0.424&    0.612&  0.058 \\
1.06 & 0.978 &0.984  &  0.988 &  0.048 &&   0.870& 0.910&    0.958&   0.052&&	0.516& 0.722&    0.856&  0.056 \\
1.08 & 1.000 &1.000  &  1.000 &  0.048 &&   0.992& 0.998&    1.000&   0.054&&	0.850& 0.934&    0.970&  0.060 \\
1.1  & 1.000 &1.000  &  1.000 &  0.066 &&   0.998& 1.000&    1.000&   0.056&&	0.968& 0.982&    0.996&  0.046 \\\hline
  \end{tabular}
\end{table}

\section{Conclusion}
\label{section:Conclusion}
We have addressed the inference problem of density ratios and its application to
homogeneity test under the semiparametric models. 
We showed that the estimator introduced by Qin \cite{Biometrika:Qin:1998} provides an
optimal estimator of the $f$-divergence with appropriate decomposition of the function
$f$, and proposed a test statistic for homogeneity test
using the optimal $f$-divergence estimator. 
It is revealed that the power function of $\widehat{D}_f$-based test does not
depend on the choice of the $f$-divergence up to the first order under the local
alternative setup. Additionally, $\widehat{D}_f$-based test and empirical
likelihood test 
\cite{fokianos01:_semip_approac_to_one_way_layout}
were shown to  
have asymptotically the same power. For misspecified density-ratio models,
we showed that
$\widehat{D}_f$-based test usually has greater power than empirical likelihood test. 
In numerical studies, mutual information based test provided the most reliable 
results than the others, that is, the null distribution was well approximated by the
asymptotic distribution with moderate samples size, and the power was comparable to 
Hotelling $T^2$-test even under the normal case. 

The choice of the $f$-divergence is an important open
problem for the homogeneity test. In our first-order asymptotic theory, the choice of the
$f$-divergence does not affect the power function. Hence, higher order asymptotic theory
may be necessary to make clear the difference among $f$-divergences for the homogeneity test.

\section{Acknowledgements}
The authors are grateful to Dr.~Hironori Fujisawa and Dr.~Masayuki Henmi of Institute of
Statistical Mathematics, and Dr.~Fumiyasu Komaki of University of Tokyo for their helpful
comments. 
T.~Kanamori was partially supported by Grant-in-Aid for Young Scientists
(20700251),
and M.~Sugiyama was supported by
SCAT, AOARD, and the JST PRESTO program.

\appendix
\section*{Appendix 1}
\subsection*{Proof of Theorem  \ref{theorem:min-variance-higher-order-expansion}}

 \begin{proof}
  Let $\delta\widehat{\theta}=\widehat{\theta}-\theta^*$. Then, due to
  \eqref{eqn:asymptotics-parameter}, we have
\begin{align*}
 \sqrt{m}\,\delta\widehat{\theta}
 &= -\sqrt{m}\,U_{\eta}^{-1}Q_{\eta}+o_p(1), 
\end{align*}
  where $\eta=\eta_{\mathrm{opt}}$ defined in \eqref{eqn:optimal-estimator}. 
  Let $\fde(r)=f(r)/(1+\rho r)$ and $\fnu(r)=\rho f(r)/(1+\rho r)$. Then
  we have 
  $\fde(1)=\fde'(1)=\fnu(1)=\fnu'(1)=0$ 
  and $\fde''(1)+\fnu''(1)=f''(1)$, 
  since $f(1)=f'(1)=0$ is assumed. 
  Hence, the asymptotic expansion of $m\bar{D}_f$ around $\theta=\theta^*$ leads to 
\begin{align*}
 m\widehat{D}_f
 &=
 \frac{\fde''(1)}{2}\sqrt{m}\delta\widehat{\theta}^T
 \Ede[\nabla{r}(x;\theta^*)\nabla{r}(x;\theta^*)^T]\sqrt{m}\delta\widehat{\theta},\\ 
 &\phantom{=}+
 \frac{\fnu''(1)}{2}\sqrt{m}\delta\widehat{\theta}^T
 \Enu[\nabla{r}(x;\theta^*)\nabla{r}(x;\theta^*)^T]\sqrt{m}\delta\widehat{\theta}+o_p(1)\\
 &= 
 \frac{(1+\rho)^2f''(1)}{2}
 \sqrt{m}Q_{\eta}^T(\Enu[\nabla{r}(x;\theta^*)\nabla{r}(x;\theta^*)^T])^{-1}
 \sqrt{m}Q_{\eta}+o_p(1), 
\end{align*}
  since $\pnu=\pde$ and $r(x;\theta^*)=1$ hold. 
  The asymptotic distribution of $\sqrt{m}Q_{\eta}$ is the Gaussian distribution 
  with mean zero and variance-covariance matrix
  $\Vnu[\nabla r]/(1+\rho)^2$, since the equality
  $\eta_{\mathrm{opt}}(x;\theta^*)=\nabla\log{r}(x;\theta^*)/(1+\rho)=\nabla{r}(x;\theta^*)/(1+\rho)$
  holds. 
  Let $M$ be the $d$ by $d$ matrix defined as
  \begin{align*}
   M=\Enu[\nabla{r}(x;\theta^*)\nabla{r}(x;\theta^*)^T], 
  \end{align*}
  and $\sqrt{V}$ be a $d$ by $d$ matrix such that 
  $\sqrt{V}\sqrt{V}^T=\Vnu[\nabla{r}]$. Then asymptotically 
  \begin{align*}
   \frac{2m}{f''(1)}\widehat{D}_f
   \ \cd \ 
   Z_d^T\sqrt{V}^T M^{-1}\sqrt{V}Z_d
  \end{align*}
  holds, where $Z_d$ is the $d$-dimensional random vector whose distribution is the
  $d$-dimensional standard Gaussian distribution, that is, $Z_d\sim N_d(0,I_d)$. 
  Let $\sqrt{M}$ be the symmetric positive definite matrix such that $M=\sqrt{M}\sqrt{M}$,
  and the vector $\mu$ be $\mu=\Enu[\nabla{r}(x;\theta^*)]$. Note that $\sqrt{M}$ is
  well-defined, since $M$ is a positive definite matrix. 
  Let $P$ be the $d$ by $d$ matrix $P=I-\sqrt{M}^{-1}\mu\mu^T\sqrt{M}^{-1}$, 
  then $P$ is the projection matrix along the vector $\sqrt{M}^{-1}\mu$. 
  Indeed, we have
  \begin{align*}
   \|\sqrt{M}^{-1}\mu\|^2=
   \Enu[\nabla{r}]^T\Enu[\nabla{r}\nabla{r}^T]^{-1}\Enu[\nabla{r}]=
   \Enu[\nabla{r}]^T b=1, 
  \end{align*}
  where $b\in\Real^d$ is the vector such that 
  $\nabla\log{r}(x;\theta^*)^Tb=\nabla r(x;\theta^*)^Tb=1$. 
  We can choose $\sqrt{V}=\sqrt{M}P$, since $\sqrt{V}\sqrt{V}^T=M-\mu\mu^T$ holds. 
  As a result, we have $Z_d^T\sqrt{V}^TM^{-1}\sqrt{V}Z_d=Z_dPZ_d$, and the distribution
  of $Z_d PZ_d$ is the chi-square distribution with $d-1$ degrees of freedom. 
 \end{proof}

\section*{Appendix 2}
\subsection*{Proof of Theorem \ref{theorem:power-function}}
First, we calculate the power function of $\widehat{D}_f$-based test. 
\begin{proof}
 Let $E[\cdot]$ be the expectation under the probability $\pde(x)=p(x)$. 
 The equality $\pnu^{(m)}(x)=\pde(x)r(x;\theta_m)$ leads to 
 $E[\nabla{r}(x;\theta^*)]^Th=0$. Indeed
 \begin{align*}
  \int\pnu^{(m)}(x)dx=\int\pde(x)r(x;\theta_m) dx
  \Longrightarrow
  1=1+E[\nabla{r}(x;\theta^*)]^T\frac{h_m}{\sqrt{m}}+o(1/\sqrt{m}) 
 \end{align*}
 holds, and thus we have $E[\nabla{r}(x;\theta^*)]^Th=0$ when $m$ tends to infinity. 
 Let $M$ be $M(\theta^*)=E[\nabla{r}(x;\theta^*)\nabla{r}(x;\theta^*)^T]$, 
 $\mu$ be $E[\nabla r(x;\theta^*)]$,  
 and $\sqrt{V}$ be a matrix such that $\sqrt{V}\sqrt{V}^T=V[\nabla r]$. 
 Let $\delta\widehat{\theta}_m$ be $\widehat{\theta}-\theta_m$. 
 Under Assumption \ref{cond:model-assumption} and Assumption
 \ref{assumption:local_alternative},  
 the asymptotic expansion provides
\begin{align*}
 &\phantom{=}\frac{2m}{f''(1)}\widehat{D}_f\\
 &=
 (\sqrt{m}\delta\theta_m+h_m)^TM (\sqrt{m}\delta\theta_m+h_m)+o_p(1)\\
 &=
 (\sqrt{m}U(\theta_m)\delta\theta_m+U(\theta_m)h)^T
 U(\theta_m)^{-1}MU(\theta_m)^{-1}
 (\sqrt{m}U(\theta_m)\delta\theta_m+U(\theta_m)h)+o_p(1)\\
 &  
 \overset{\theta_m}{\longrightarrow} \ 
 \big\|\sqrt{M}^{-1}\sqrt{V}Z_d +\sqrt{M}h \big\|^2. 
\end{align*}
 In the same way as the proof of Theorem
 \ref{theorem:min-variance-higher-order-expansion}, 
 we see that $\sqrt{M}^{-1}\sqrt{V}$ is the projection matrix along the vector $\sqrt{M}^{-1}\mu$. 
 Moreover, $\sqrt{M}h$ is orthogonal to the vector $\sqrt{M}^{-1}\mu$ since
 $\mu^Th=0$ holds. As a result, we see that the distribution function of 
 $\big\|\sqrt{M}^{-1}\sqrt{V}Z_d +\sqrt{M}h \big\|^2$
 is the non-central chi-square distribution with $d-1$ degrees of freedom and
 non-centrality parameter $h^TM(\theta^*)h$. 
\end{proof}

Next, we calculate the power function of empirical likelihood test. 
The notations $M$ and $\mu$ are the same as the proof above. 
\begin{proof}
 From the definition of the statistic $S$, we have
 \begin{align*}
  S=m(\widehat{\beta}-\beta^*)^T\Vnuh[\nabla_\beta\phi](\widehat{\beta}-\beta^*)
  =m
  (\widehat{\theta}-\theta^*)^T V(\widehat{\theta}-\theta^*)+o_p(1), 
 \end{align*}
 where $V=V[\nabla{r}]$. Then we have
 \begin{align*}
  &\phantom{=}
  m (\widehat{\theta}-\theta^*)^T V(\widehat{\theta}-\theta^*)+o_p(1)\\
  &=
 (\sqrt{m}U(\theta_m)\delta\theta_m+U(\theta_m)h)^T
 U(\theta_m)^{-1}VU(\theta_m)^{-1}
 (\sqrt{m}U(\theta_m)\delta\theta_m+U(\theta_m)h)+o_p(1)\\
 &  
 \overset{\theta_m}{\longrightarrow} \ 
 \big\|\sqrt{V}^T\sqrt{M}^{-1}(\sqrt{M}^{-1}\sqrt{V}Z_d +\sqrt{M}h) \big\|^2. 
 \end{align*}
 The matrix $\sqrt{V}^T\sqrt{M}^{-1}$ is the projection matrix along the vector
 $\sqrt{M}^{-1}\mu$ and $\mu^Th=0$ holds. Then we see that the vector 
 $\sqrt{M}^{-1}\sqrt{V}Z_d +\sqrt{M}h$ is orthogonal to $\sqrt{M}^{-1}\mu$. 
 This implies 
 \begin{align*}
  \big\|\sqrt{V}^T\sqrt{M}^{-1}(\sqrt{M}^{-1}\sqrt{V}Z_d +\sqrt{M}h) \big\|^2
  =
  \big\|\sqrt{M}^{-1}\sqrt{V}Z_d +\sqrt{M}h\big\|^2. 
 \end{align*}
 Thus, under the local alternative setup, the limit distribution of the test statistic $S$ 
 is the non-central chi-square distribution with the same parameter as
 $\widehat{D}_f$-based test. 
\end{proof}

\section*{Appendix 3}
\subsection*{Proof of Theorem \ref{theorem:power_misspecifiedmodel}}

Below, the notations $M=E[\nabla r(x;\theta^*)\nabla r(x;\theta^*)]$ and
$\mu=E[\nabla{r}(x;\theta^*)]$ are used. 
\begin{proof}
From the definition of the density $\pnu^{(m)}(x)$, we have
 \begin{align*}
  &\phantom{\Longrightarrow}
  \int\pnu^{(m)}(x)dx=\int\pde(x)
  \left(r(x;\theta_m)+\frac{s_m(x)+\varepsilon_m}{\sqrt{\mde}}\right) dx\\
  &\Longrightarrow
  1=1+E[\nabla{r}(x;\theta^*)]^T\frac{h_{m}}{\sqrt{m}}+
  \frac{\varepsilon_m}{\sqrt{m}}+o(1/\sqrt{m}), 
 \end{align*}
 and thus, the equality $\mu^T h+\varepsilon=0$ holds when $m$ tends to infinity. 
 Let the random vector $W$ be
 \begin{align*}
  W=PZ_d+\sqrt{M}h, \qquad Z_d\sim N_d(0,I_d), 
 \end{align*}
 where $P$ is the projection matrix along the vector $\sqrt{M}^{-1}\mu$
 as defined in the proof of Theorem \ref{theorem:min-variance-higher-order-expansion}. 
 According to the proof in Theorem \ref{theorem:power-function} in Appendix 2. 
the power of $\widehat{D}_f$-based 
 test is asymptotically equal to 
 $\Pr\big\{  \|W\|^2\geq \chi^2_{d-1}(1-\alpha)  \big\}$, 
 and that of empirical likelihood test is equal to 
 $\Pr\big\{  \|PW\|^2\geq \chi^2_{d-1}(1-\alpha)  \big\}$. 
 We have the equality 
 $W=PW+c\sqrt{M}^{-1}h$ with some $c\in\Real$. 
 Note that generally $\sqrt{M}h$ is not orthogonal to $\sqrt{M}^{-1}\mu$ in the misspecified
 case, since 
 \begin{align*}
  (\sqrt{M}^{-1}\mu)^T\sqrt{M}h=\mu^Th=-\varepsilon
 \end{align*}
 holds. 
 For $\varepsilon\neq0$, we have $c\neq 0$ and then the inequality $\|W\|^2>\|PW\|^2$
 holds. As a result, the power of $\widehat{D}_f$-based test is larger than or equal to
 that of empirical likelihood test under the misspecified setup. 
\end{proof}

\bibliographystyle{plain}

\end{document}